\pdfoutput=1
\PassOptionsToPackage{table}{xcolor}
\documentclass[11pt]{article}

\usepackage[]{naacl2021}

\usepackage{times}
\usepackage{latexsym}
\usepackage{graphicx}
\usepackage{amsmath}
\usepackage{amsthm}
\usepackage{amsfonts}
\usepackage{amssymb}
\usepackage{amscd}

\usepackage[T1]{fontenc}

\usepackage[utf8]{inputenc}

\usepackage{microtype}

\newcommand{\Real}{\mathbb{R}}

\title{Leveraging Dependency Grammar for Fine-Grained Offensive Language Detection using Graph Convolutional Networks}

\author{Divyam Goel \\
  Indian Institute of Technology Roorkee  \\
  Roorkee, India \\
  \texttt{dgoel@bt.iitr.ac.in} \\\And
  Raksha Sharma \\
  Indian Institute of Technology Roorkee \\
  Roorkee, India \\
  \texttt{raksha.sharma@cs.iitr.ac.in} \\}

\begin{document}
\maketitle
\begin{abstract}
The last few years have witnessed an exponential rise in the propagation of offensive text on social media. 
Identification of this text with high precision is crucial for the well-being of society. 
Most of the existing approaches tend to give high toxicity scores to innocuous statements (\emph{e.g.}, ``I am a gay man''). 
These false positives result from over-generalization on the training data where specific terms in the statement may have been used in a pejorative sense (\emph{e.g.}, ``gay''). 
Emphasis on such words alone can lead to discrimination against the classes these systems are designed to protect. 
In this paper, we address the problem of offensive language detection on Twitter, while also detecting the type and the target of the offense. 
We propose a novel approach called \emph{SyLSTM}, which integrates syntactic features in the form of the dependency parse tree of a sentence and semantic features in the form of word embeddings into a deep learning architecture using a Graph Convolutional Network. 
Results show that the proposed approach significantly outperforms the state-of-the-art BERT model with orders of magnitude fewer number of~parameters.
\end{abstract}

\section{Introduction}
Offensive language can be defined as instances of profanity in communication, or any instances that disparage a person or a group based on some characteristic such as race, color, ethnicity, gender, sexual orientation, nationality, religion, \emph{etc.}~\cite{nockleby:2000}. 
The ease of accessing social networking sites has resulted in an unprecedented rise of offensive content on social media. 
With massive amounts of data being generated each minute, it is imperative to develop scalable systems that can automatically filter offensive content. 

The first works in offensive language detection were primarily based on a lexical approach, utilizing surface-level features such as n-grams, bag-of-words, \emph{etc.}, drawn from the similarity of the task to another NLP task, {\em i.e.}, Sentiment Analysis (SA). 
These systems perform well in the context of foul language but prove ineffective in detecting hate speech. 
Consequently, the main challenge lies in discriminating profanity and hate speech from each other~\cite{zampieri:2019}. 
On the other hand, recent deep neural network based approaches for offensive language detection fall prey to inherent biases in a dataset, leading to the systems being discriminative against the very classes they aim to protect. 
Davidson et al., \shortcite{davidson:2019} presented the evidence of a systemic bias in classifiers, showing that such classifiers predicted tweets written in African-American English as abusive at substantially higher rates. 
Table \ref{tab:table6} presents the scenarios where a tweet may be considered~hateful.
\begin{table*}[hbtp!]
    \centering
    \begin{tabular}{|c|l|}
    \hline
      \textbf{S.No.} & \textbf{Hateful Tweet Scenarios}  \\
    \hline
       1 & uses sexist or racial slurs.   \\
       2 & attacks a minority.  \\
       3 & seeks to silence a minority. \\
       4 & criticizes a minority (without a well-founded argument). \\
       5 & promotes but does not directly use hate speech or violent crime. \\
       6 & criticizes a minority and uses a straw man argument. \\
       7 & blatantly misrepresents truth or seeks to distort views on a minority with unfounded claims.\\
       8 & shows support of problematic hashtags. {\em E.g.} “\#BanIslam,” “\#whoriental,” “\#whitegenocide”\\
       9 & negatively stereotypes a minority.\\
       10 & defends xenophobia or sexism.\\
       11 & contains an offensive screen name\\
      \hline
    \end{tabular}
    \caption{Hateful Tweet Scenarios~\cite{waseem:2016}}
    \label{tab:table6}
    \vspace{-0.5cm}
\end{table*}

Syntactic features are essential for a model to detect latent offenses, {\em i.e.}, untargeted offenses, or where the user might mask the offense using the medium of sarcasm~\cite{schmidt:2017}. 
Syntactic features prevent over-generalization on specific word classes, {\em e.g.}, profanities, racial terms, \emph{etc.}, instead examining the possible arrangements of the precise lexical internal features which factor in differences between words of the same class. 
Hence, syntactic features can overcome the systemic bias, which may have arisen from the pejorative use of specific word classes. 
A significant property of dependency parse trees is their ability to deal with morphologically rich languages with a relatively free word order~\cite{10.5555/1214993}. 
Motivated by the nature of the modern Twitter vocabulary, which also follows a relatively free word order, we present an integration of syntactic features in the form of dependency grammar in a deep learning framework.

In this paper, we propose a novel architecture called \emph{Syntax-based LSTM} (\emph{SyLSTM}), which integrates latent features such as syntactic dependencies into a deep learning model. 
Hence, improving the efficiency of identifying offenses and their targets while reducing the systemic bias caused by lexical features. 
To incorporate the dependency grammar in a deep learning framework, we utilize the Graph Convolutional Network (GCN)~\cite{kipf:2016}. 
We show that by subsuming only a few changes to the dependency parse trees, they can be transformed into compatible input graphs for the GCN. 
The final model consists of two major components, a BiLSTM based Semantic Encoder and a GCN-based Syntactic Encoder in that order. 
Further, a Multilayer Perceptron handles the classification task with a Softmax head.
The state-of-the-art BERT model requires the re-training of over $110M$ parameters when fine-tuning for a downstream task. 
In comparison, the \emph{SyLSTM} requires only $\sim9.5M$ parameters and significantly surpasses BERT level performance. 
Hence, our approach establishes a new state-of-the-art result for offensive language detection while being over ten times more parameter efficient than BERT.

We evaluate our model on two datasets; one treats the task of hate speech and offensive language detection separately~\cite{davidson:2017}.
The other uses a hierarchical classification system that identifies the types and targets of the offensive tweets as a separate task~\cite{zampieri:2019}. 

\paragraph{Our Contribution:} The major contribution of this paper is to incorporate syntactic features in the form of dependency parse trees along with semantic features in the form of feature embeddings into a deep learning architecture. 
By laying particular emphasis on sentence construction and dependency grammar, we improve the performance of automated systems in detecting hate speech and offensive language instances, differentiating between the two, and identifying the targets for the same. 
Results (Section \ref{result}) show that our approach significantly outperforms all the baselines for the three tasks, {\em viz.}, identification of offensive language, the type of the offense, and the target of the offense.

The rest of the paper is organized as follows. 
In Section \ref{related}, we discuss related work in this field. Section \ref{methodology} presents the design of \emph{SyLSTM}. 
Section \ref{experimental} elaborates on the datasets and the experimental protocol.
Section \ref{result} presents the results and discussion, and Section \ref{conclusion} concludes the paper.

\section{Related Work}
\label{related}
Hate speech detection, as a topical research problem, has been around for over two decades. 
One of the first systems to emerge from this research was called {\em Smokey}~\cite{spertus:1997}. 
It is a decision-tree-based classifier that uses $47$ syntactic and semantically essential features to classify inputs in one of the three classes ($ flame $, $ okay $ or $ maybe $). 
\emph{Smokey} paved the way for further research in using classical machine learning techniques to exploit the inherent features of Natural Language over a plethora of tasks such as junk filtering~\cite{sahami:1998}, opinion mining~\cite{wiebe:2005}~\emph{etc}.

Owing to the unprecedented rise of social networks such as Facebook and Twitter, most of the research on hate speech detection has migrated towards the social media domain. 
To formalize this new task, a set of essential linguistic features was proposed~\cite{waseem:2016}. 
Initial research in this direction focused more on detecting profanity, pursuing hate speech detection implicitly~\cite{nobata:2016,waseem:2017}. 
Using these systems, trained for detecting profanities, to detect hate speech reveals that they fall prey to inherent biases in the datasets while also proving ineffective in classifying a plethora of instances of hate~speech~\cite{davidson:2019}.

Research has also shown the importance of syntactic features in detecting offensive posts and identifying the targets of such instances~\cite{chen:2012}. 
On social media, it was found that hate speech is primarily directed towards specific groups, targeting their ethnicity, race, gender, caste, {\em etc.}~\cite{silva:2016}.
ElSherief et al. \shortcite{elsherief:2018} make use of linguistic features in deep learning models, which can be used to focus on these directed instances. 
The problem with this approach is two-fold. 
First, these linguistic features learn inherent biases within the datasets, thus discriminating against the classes they are designed to protect. 
Second, the use of explicit linguistic features to detect hate speech leaves the model prone to the effects of domain shift.
Altogether, there is a need to develop more robust techniques for hate speech detection to address the above mentioned issues.
While the use of syntactic features for the task has proven useful, there has been little effort towards incorporating non-Euclidean syntactic linguistic structures such as dependency trees into the deep learning~sphere.

Graph Neural Networks (GNNs) provide a natural extension to deep learning methods in dealing with such graph structured data. 
A special class of GNNs, known as Graph Convolutional Networks (GCNs), generalize Convolutional Neural Networks (CNNs) to non-Euclidean data. 
The GCNs were first introduced by Bruna et al. \shortcite{bruna2013spectral}, following which, Kipf et al. \shortcite{kipf:2016} presented a scalable,  first order approximation of the GCNs based on Chebyshev polynomials. 
The GCNs have been extremely successful in several domains such as social networks~\cite{hamilton2017inductive}, natural language processing~\cite{marcheggiani2017encoding} and natural sciences~\cite{zitnik2018modeling}. 

Marcheggiani and Titov \shortcite{marcheggiani2017encoding} were the first to show the effectiveness of GCNs for NLP by presenting an analysis over semantic role labelling.
Their experiments paved the way for researchers to utilize GCNs for feature extraction in NLP. 
Since then, GCNs have been used to generate embedding spaces for words~\cite{vashishth:2018}, documents~\cite{peng:2018} and both words and documents together~\cite{yao:2019}. 
Even though GCNs have been used in NLP, their inability to handle multirelational graphs has prevented researchers from incorporating the dependency parse tree in the deep feature space. 

In this paper, we present a first approach towards transforming the dependency parse tree in a manner that allows the GCN to process it. 
The final model is a combination of a BiLSTM based Semantic Encoder, which extracts semantic features and addresses long-range dependencies, and a GCN-based Syntactic Encoder, which extracts features from the dependency parse tree of the sentence. 
Results show that the proposed approach improves the performance of automated systems in detecting hate speech and offensive language instances, differentiating between the two, and identifying the targets for the same.

\section{Methodology}
\label{methodology}
Traditionally, grammar is organized along two main dimensions: \emph{morphology} and \emph{syntax}. 
While morphology helps linguists understand the structure of a word, the syntax looks at sentences and how each word performs in a sentence. 
The meaning of a sentence in any language depends on the syntax and order of the words. 
In this regard, a sentence that records the occurrence of relevant nouns and verbs ({\em e.g.}, Jews and kill) can prove helpful in learning the offensive posts and their targets~\cite{gitari:2015}. 
Further, the syntactic structure  I $\langle intensity \rangle$ $\langle user intent \rangle$ $\langle hate target \rangle$, {\em e.g.}, “I f*cking hate white people,” helps to learn more about offensive posts, their targets, and the intensity of the offense~\cite{silva:2016}. 
Our approach incorporates both semantic features and the dependency grammar of a tweet into the deep feature space. 
The following subsections present a detailed discussion on the proposed methodology.
\begin{table*}[hbt!]
    \centering
    \resizebox{1.8\columnwidth}{!}{
    \begin{tabular}{|c|c|}
    \hline
      \textbf{Preprocessing} & \textbf{Description}  \\
    \hline
      {\em Replacing usernames} & replacing all usernames with ‘@user’. {\em Eg.} ‘@india’ to ‘@user’.   \\
      {\em Replacing URLs} & replacing URLs in a tweet with the word ‘url’.  \\
      {\em Hashtag Segmentation} &  Eg. ‘\#banislam’ becomes ‘\# banislam’. \\
      {\em Emoji Normalization} & normalizing emoji instances with text. {\em Eg.} ‘:)’ becomes ‘smiley face’. \\
      {\em Compound Word Splitting} & split compound words. {\em E.g.} ‘putuporshutup’ to ‘put up or shut up’. \\
      {\em Reducing Word Lengths} & reduce word lengths, exclamation marks, {\em E.g.} ‘waaaaayyyy’ to ‘waayy’. \\
      \hline
    \end{tabular}
    }
    \caption{Preprocessing Modules}
    \label{tab:table5}
    \vspace{-0.3cm}
\end{table*}
\begin{figure*}[hbt!]
    \centering
    \includegraphics[width=10cm,height=8cm]{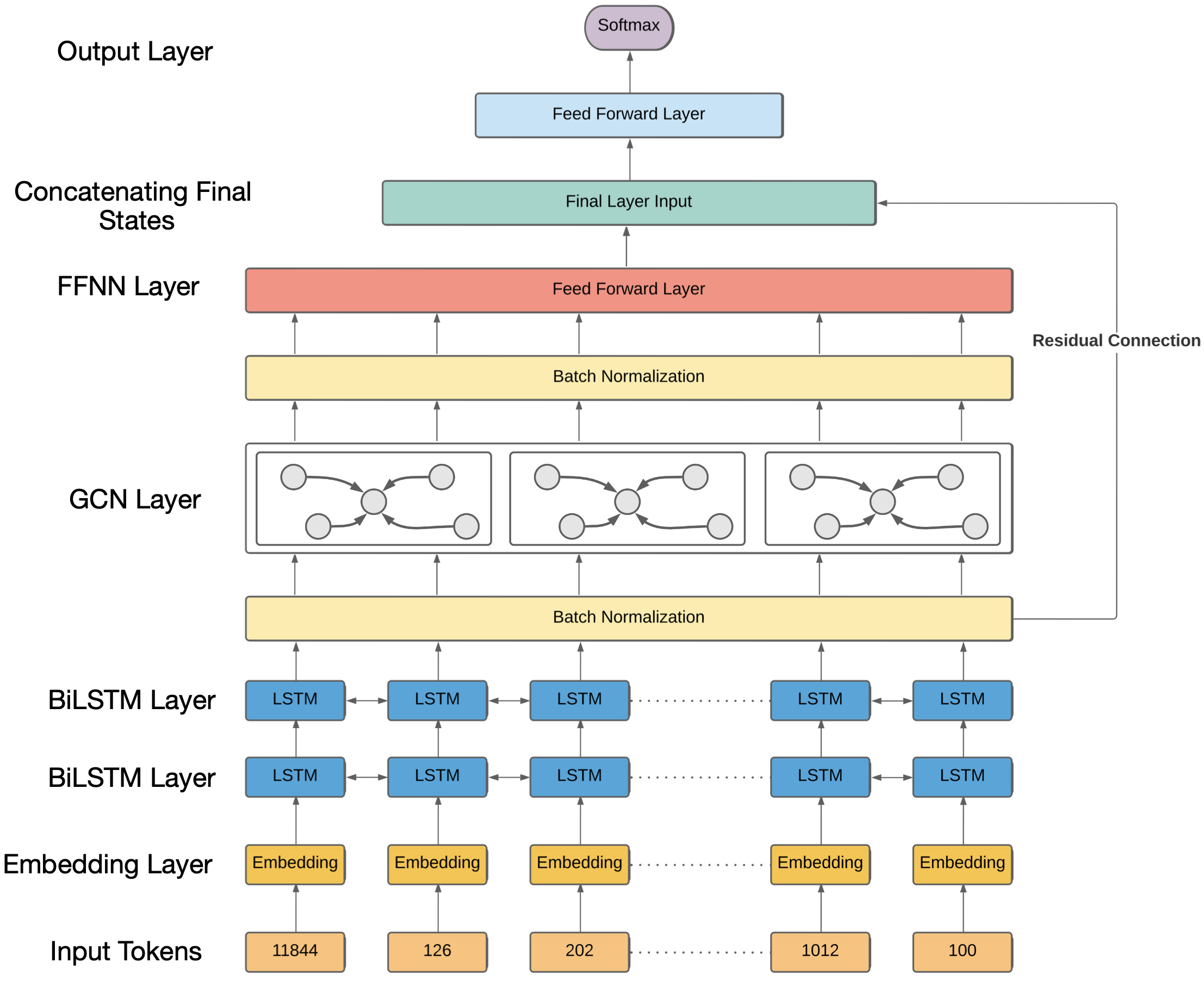}
    \caption{Model Architecture for \emph{SyLSTM}}
    \label{fig:model}
    \vspace{-0.5cm}
\end{figure*}

\subsection{Preprocessing}
Raw tweets usually have a high level of redundancy and noise associated with them, such as varying usernames, URLs, \emph{etc}. 
In order to clean the data, we implement the preprocessing module described in Table \ref{tab:table5}.

\subsection{Model}
The proposed model \emph{SyLSTM} (Figure \ref{fig:model}) has the following six components:
\begin{enumerate}
    \item Input Tokens: The tweet is passed through a word-based tokenizer after the preprocessing step. The tokenized tweet is then given as input to the model;
    \item Embedding Layer: A mapping for each word to a low-dimensional feature vector;
    \item BiLSTM Layer: used to extract a high-level feature space from the word embeddings;
    \item GCN Layer: produces a weight vector according to the syntactic dependencies over the high-level features from step $3$. 
    Multiply with high-level feature space to produce new features with relevant syntactic information.
    \item Feed Forward Network: reduces the dimensionality of the outputs of step $4$.
    \item Output Layer: the last hidden states from step $3$ are concatenated with the output of step 5 as a residual connection and fed as input. 
    The feature space is finally used for hate speech detection.
\end{enumerate}
The detailed description of these components is given below.

\paragraph{Word Embeddings:}  Given a sentence consisting of $ T $ words $S = \{x_1, x_2, ... , x_T\}$, every word $x_i$ is converted to a real valued feature vector $e_i$.
This is done by means of an embedding matrix which serves as a lookup table, 
\begin{equation}
    \label{eq:embedding_matrix}
        \mathcal{E}^{(word)} \in \Real^{|V| \times d^{(w)}},
\end{equation}
where, $|V|$ is the size of the vocabulary and $d^{(w)}$ is the dimensional size of the embeddings. 
Each word in $ S $ is then mapped to a specific entry in this~matrix, 
\begin{equation}
    \label{eq:embedding}
        e_i = \mathcal{E}^{(word)} . v_i,
\end{equation}
where, $v_i$ is a one hot vector of size $|V|$. 
The entire sentence is fed into the proceeding layers as real-valued vectors $emb = \{e_1, e_2, ... , e_T\}$. 
The embedding matrix can be initialized randomly and learned via backpropagation, or one can also use a set of pretrained embeddings. 
Twitter posts generally use the modern \emph{internet lexicon} and hence have a unique vocabulary. 
For our model, we use two different instances for the embedding space - first, a randomly initialized embedding space learned at the training time. 
Second, a pretrained embedding space where we utilize the GloVe-Twitter Embeddings\footnotemark[1]\footnotetext[1]{\url{https://nlp.stanford.edu/projects/glove/}} $(d^{(w)} = 200)$. 
These embeddings have been trained on 27B tokens parsed from a Twitter corpus~\cite{pennington:2014}. 
Results indicate that models trained on the GloVe-Twitter Embeddings learn a stronger approximation of semantic relations in the twitter vocabulary, showcasing a more robust performance than their randomly initialized~counterparts.

\paragraph{Semantic Encoding with BiLSTM:} Most of the existing research on GCNs focuses on learning nodal representations in undirected graphs. 
These are suited to single relational edges and can suffer from a severe semantic gap when operating on multirelational graphs. 
To codify the relational edges' underlying semantics and resolve language on a temporal scale, we utilize the Bidirectional LSTM.

Using an adaptive gating mechanism, the LSTMs decide the degree of importance between features extracted at a previous time step to that at the current time step~\cite{hochreiter:1997}. 
Consequently, they prove extremely useful in the context of hate speech detection, where hate speech can be distributed randomly at any part of the sentence. 
Standard LSTMs process sequences in a temporal order hence ignoring future context. 
Bidirectionality allows us access to both future and past contexts, which helps improve the cognition of hate speech in a tweet~\cite{8684825}. 

We pass the sentence embedding vectors $emb = \{e_1, e_2, ... , e_T\}$ through a two-layered BiLSTM network with $32$ hidden units and a dropout of $0.4$. 
As outputs, we extract the sequential vectors and the final hidden states for the forward and backward sequences. 
The final hidden states for the forward and backward sequences are concatenated and used as a residual connection at a later stage, as shown in Figure \ref{fig:model}. 
The sequential vectors are passed through a batch normalization layer with a momentum of $0.6$ and then fed into the GCN layer along with the dependency parse trees.
\begin{figure*}[hbt!]
    \centering
    \includegraphics[width=13cm,height=4cm]{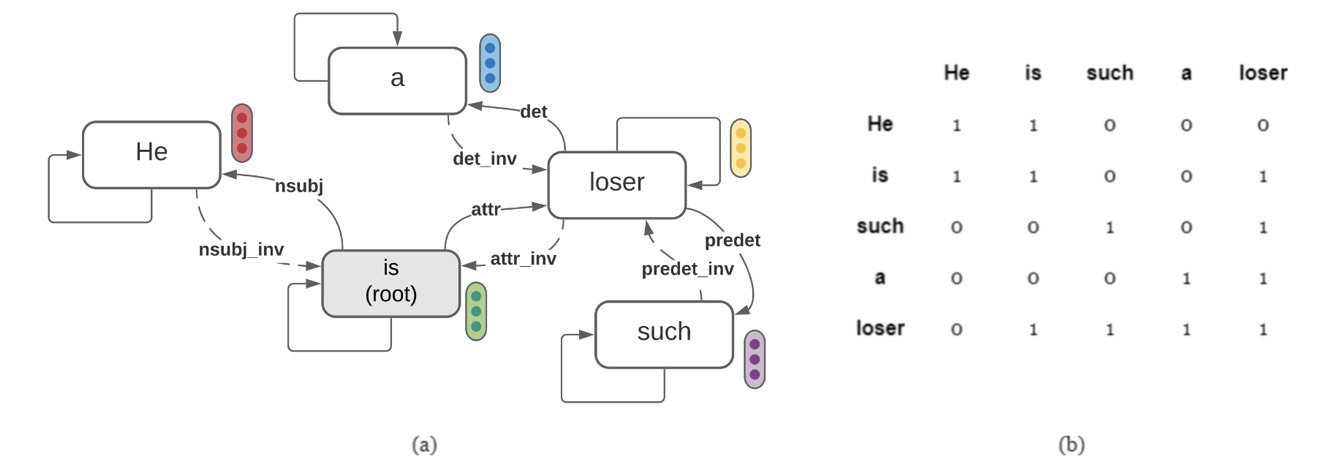}
    \caption{(a) Dependency Graph $ G $ with Nodal Embeddings (b) Adjacency Matrix $ A $ for the graph $ G $}
    \label{fig:graph}
    \vspace{-0.5cm}
\end{figure*}

\paragraph{Syntactic Encoding with GCN:}  The dependency parse trees have specific characteristics which are rarely considered in general graphs. 
On the one hand, they have multirelational edges. 
And on the other hand, the definition of each type of edge is relatively broad, resulting in a huge difference in the semantics of edges with the same relationship. 
For instance, an `amod' dependency may be presented in <Techniques, Computational> and <Techniques, Designed>, but their semantics are obviously different.

The GCN~\cite{kipf:2016} cannot handle such scenarios without introducing some changes to the structure of the input dependency parse tree. 
First, inverse edges corresponding to each of the respective dependencies are introduced between all connected nodes. 
Furthermore, to highlight the importance of specific words in the given context, we add self-loops over each node. 
The dependency parse tree of a sentence is extracted using the NLP open-source package spaCy\footnotemark[2]\footnotetext[2]{\url{https://github.com/explosion/spaCy}}. 

Hence, the extracted dependency parse tree is transformed into a graph $G = (V, E)$, where $V$ is the set of all vertices which represent the words in a tweet and $E$ is the set of all edges which highlight the dependency and their inverse relations. 
The result is an undirected graph with self-loops (see Figure \ref{fig:graph}). 
This comes as a natural extension to the dependency structure of the sentence, highlighting the importance of word positioning and combating possible confusions in identifying the direction of the dependency. 
The graph is then fed into the GCN as a sparse adjacency matrix, with each dependency represented by a weight $\alpha$. 
With the setup in place, the GCN performs a convolution operation over the graph $G$ represented by the adjacency matrix $A$. 
Formally, the GCN performs the following~computation:
\begin{equation}
    \label{eq:GCN}
        H^{(l+1)} = \sigma(\tilde D^{-\frac{1}{2}} \tilde A \tilde D^{-\frac{1}{2}}H^{(l)}W^{(l)})
\end{equation}
where, $\tilde A = A + I_N$ is the adjacency matrix of the undirected graph $ G $ with added self-connections. 
$I_N$ is the identity matrix, $\tilde D_{ii} = \Sigma_j \tilde A_{ij}$ and $W^{(l)}$ is a layer-specific trainable weight matrix. 
$\sigma (\cdot)$ denotes an activation function, in our case the ReLU$(\cdot)$ = $\max(0, \cdot)$. 
$H^{(l)}  \in  \Real^{N \times D}$ is the matrix of activations in the $l^{th}$ layer; $H^{(0)} = L$. 
The model learns hidden layer representations that encode both local graph structure (\emph{the dependencies}) and nodal features (\emph{the importance of the word in that context}). 
Furthermore, the Semantic Encoder complements the Syntactic Encoder by addressing the long range spatial inabilities of the GCN~\cite{marcheggiani2017encoding}. 
The sparse adjacency matrix leads to a problem with vanishing gradients. 
We combat this by applying a batch normalization layer with a momentum of $0.6$ and applying a dropout of $0.5$. 
We use the Xavier distribution to initialize the weight matrix and set the output dimension of the GCN as $32$.

\paragraph{Feed Forward Neural Network (FFNN):}  The output of the GCN is then passed through a single layered FFNN to learn high-level features based on dependency structure.
The FFNN is activated using the non-linear ReLU activation function.

\paragraph{Output Layer:} The output from the FFNN is then concatenated with the last hidden states of the BiLSTM which is added as a residual connection. 
The concatenated vector is then passed through a linear layer with a softmax head that produces a probability distribution over the required outputs. 

\section{Experimental Setup}
\label{experimental}
This section describes the dataset and the experimental setup for the models reported in the paper. 
\subsection{Datasets}
The primary motivation of this paper is the design of a methodology to integrate a neural network model with syntactic dependencies for improved performance over fine-grained offensive language detection. 
Keeping in line with this ideology, we test our model on two separate datasets.
The following section describes these datasets at length.

\paragraph{Offensive Language Identification Dataset:}  
This dataset was presented for a shared task on offensive language detection in the SemEval Challenge $2019$. 
This was the first time that offensive language identification was presented as a hierarchical task. 
Data quality was ensured by selecting only experienced annotators and using test questions to eliminate individuals below a minimum reliability threshold. 
Tweets were retrieved using a keyword approach on the Twitter API. 
The dataset forms a collection of $14,000$ English tweets annotated for three subtasks proceeding in a hierarchy~\cite{zampieri:2019}:
\begin{enumerate}
    \item whether a tweet is offensive or not (A);
    \item whether the offensive tweet is targeted (B);
    \item whether the target of the offensive tweet is an individual, a group, or other (\emph{i.e.}, an organization, an event, an issue, a situation) (C).
\end{enumerate}
We choose this dataset because of the extended subtasks B and C. 
An increase in performance over these will posit that our model has been successful in tackling its objectives. 
We evaluate our model on all three subtasks.

\paragraph{Hate Speech and Offensive Language Dataset:}  Motivated by the central problem surrounding the separation of hate speech from other instances of offensive language, Davidson et al. \shortcite{davidson:2017} curated a dataset annotating each tweet in one of three classes, hate speech (\textbf{HATE}), offensive language (\textbf{OFF}), and none (\textbf{NONE}). 
They use a hate speech lexicon containing words and phrases identified by internet users as hate speech, compiled by \emph{Hatebase}. 
These lexicons are used to extract English tweets from the Twitter API. 
From this corpus, a random sample of $25k$ tweets containing terms from the lexicon was extracted. 
The tweets were manually coded by CrowdFlower (CF) workers, with a final inter-annotator agreement of $92\%$. 
\subsection{Baseline Models}
In the following section, we describe the design of all the baseline models used for comparison.

\paragraph{Linear-SVM:} SVMs have achieved state-of-the-art results for many text classification tasks and significantly outperform many neural networks over the OLID dataset~\cite{zampieri:2019}. 
Hence, we use a Linear-SVM trained on word unigrams as a baseline.  
We employ a Grid-search technique to identify the best hyperparameters. 

\paragraph{Two-channel BiLSTM:}  We design a two-channel BiLSTM as a second baseline, with the two input channels differentiated only by their embedding space. 
One of the input channels learns the embedding space via backpropagation after a random initialization, while the other uses the pretrained BERT embeddings. 
This choice is motivated by the contextual nature of the BERT embeddings. 
This conforms with the ideation that certain words may be deemed offensive depending upon the context they are used in. 
The BiLSTM itself is two layers deep and consists of $32$ hidden-units. 
The final hidden states for the forward and backward sequences of each channel are concatenated and passed through an MLP with a softmax head for classification. 

\paragraph{Fine-tuned BERT:}  We also fine-tune a BERT model~\cite{devlin:2018} for this task. 
We adapt the state-of-the-art BERT model which won the SemEval Challenge $2019$~\cite{liu:2019} and tune the hyperparameters of the model to get the best performance on our preprocessing strategy. 
While fine-tuning this model, the choices over the loss function, optimizer, and learning rate schedule remain the same as those for the \emph{SyLSTM}. 

\subsection{Training} We train our models using the standard cross-entropy loss.
The AdamW optimizer~\cite{loshchilov:2018} is chosen to learn the parameters. 
To improve the training time and chances of reaching the optima, we adopt a cosine annealing~\cite{loshchilov:2017} learning rate scheduler.
The vocabulary of the models is fixed to the top $30,000$ words in the corpus. The initial learning rate is set to $0.001$, with a regularization parameter of $0.1$. 

\subsection{Evaluation Metric}
The datasets exhibit large class imbalances over each task. 
In order to address this problem, we use the Weighted F1-measure as the evaluation metric. 
We also provide the precision and recall scores for a deeper insight into the model’s performance.

\section{Results}
\label{result}
We evaluate two instances of our model, (1) with a randomly initialized embedding matrix (referred to as \emph{SyLSTM}) and (2) utilizing the pretrained GloVe Twitter embeddings (referred to as \emph{SyLSTM*}). 
A paired Student's t-test using the Weighted-F1 measure of the model's performance shows that our models significantly outperform each of the baselines across all the tasks (\emph{p $<$ 0.001}).

\subsection{Performance on Offensive Language Identification Dataset}
In this section, we present performance comparisons between the baselines and the \emph{SyLSTM} for the three subtasks. 
We split the training data, using $10\%$ of the tweets to get a dev set. 
The hyperparameters are tuned according to the performance on the dev set. 
The results presented here demonstrate the performance over the predefined test set. 
We also present the performance metrics for the trivial case, notably where the model predicts only a single label for each tweet. 
By comparison, we show that the chosen baselines and our models perform significantly better than chance for each~task. 

\paragraph{Offensive Language Detection:} The performance comparisons for discriminating between offensive (\textbf{OFF}) and non-offensive (\textbf{NOT}) tweets are reported in Table \ref{tab:table1}. 
Neural network models perform substantially better than the Linear-SVM. 
Our model (in gray) outperforms each of the baselines in this task.
\begin{table}[hbt!]
    \centering
    \resizebox{0.9\columnwidth}{!}{
    \begin{tabular}{l l l l}
    \hline
     \textbf{System} & \textbf{Precision} & \textbf{Recall} & \textbf{F1-score} \\
     \hline
      All OFF & 8.4 & 28.2 & 12.1 \\
      All NOT & 52.4 & 72.7 & 60.4 \\
    \hline
      SVM & 77.7 & 80.2 & 78.6\\
      BiLSTM & 81.7 & 82.8 & 82.0\\
      BERT & \underline{87.3} & 85.8 & 85.7\\
    \hline
      \rowcolor[RGB]{230,230,230}
      SyLSTM & 85.2 & \underline{88.1} & \underline{86.4}\\
      \rowcolor[RGB]{230,230,230}
      SyLSTM* & \textbf{87.6} & \textbf{88.1} & \textbf{87.4}\\
      \hline
    \end{tabular}
    }
    \caption{Offensive Language Detection}
    \label{tab:table1}
    \vspace{-0.5cm}
\end{table}

\paragraph{Categorization of Offensive Language:}  This sub-task is designed to discriminate between targeted insults and threats (\textbf{TIN}) and untargeted (\textbf{UNT}) offenses, generally referring to profanity~\cite{zampieri:2019}.
Performance comparisons for the same are reported in Table 4. 
Our model (in gray) shows a significant $4\%$ relative improvement in performance in comparison to the BERT model.
\begin{table}[hbt!]
    \centering
    \resizebox{0.9\columnwidth}{!}{
    \begin{tabular}{l l l l}
    \hline
     \textbf{System} & \textbf{Precision} & \textbf{Recall} & \textbf{F1-score} \\
     \hline
      All TIN & 78.7 & 88.6 & 83.4 \\
      All UNT & 1.4 & 11.3 & 12.1 \\
    \hline
      SVM & 81.6 & 84.1 & 82.6\\
      BiLSTM & 84.8 & 88.4 & 85.7\\
      BERT & 88.4 & \underline{92.3} & 89.6\\
    \hline
      \rowcolor[RGB]{230,230,230}
      SyLSTM & \underline{90.6} & 91.6 & \underline{91.4}\\
      \rowcolor[RGB]{230,230,230}
      SyLSTM* & \textbf{94.4} & \textbf{92.3} & \textbf{93.2}\\
      \hline
    \end{tabular}
    }
    \caption{Categorization of Offensive Language}
    \label{tab:table2}
    \vspace{-0.5cm}
\end{table}

 \paragraph{Offensive Language Target Identification:}  This sub-task is designed to discriminate between three possible targets: a group (\textbf{GRP}), an individual (\textbf{IND}), or others (\textbf{OTH}). 
 The results for the same are reported in Table \ref{tab:table3}. 
 Note that the three baselines produce almost identical results. 
 The low F1-scores for this task may be on account of the small size of the dataset and large class imbalances, factors that make it difficult to learn the best features for classification. 
 Our model (in gray) shows a $5.7\%$ relative improvement over the BERT model, hence showcasing its robustness when generalizing over smaller datasets.
\begin{table}[hbt!]
    \centering
    \resizebox{0.9\columnwidth}{!}{
    \begin{tabular}{l l l l}
    \hline
     \textbf{System} & \textbf{Precision} & \textbf{Recall} & \textbf{F1-score} \\
     \hline
      All GRP & 13.6 & 37.4 & 19.7 \\
      All IND & 22.1 & 47.3 & 30.3 \\
      ALL OTH & 3.4 & 16.2 & 5.4 \\
    \hline
      SVM & 56.1 & 62.4 & 58.3\\
      BiLSTM & 56.1 & 65.8 & 60.4\\
      BERT & 58.4 & 66.2 & 60.9\\
    \hline
      \rowcolor[RGB]{230,230,230}
      SyLSTM & \underline{60.3} & \textbf{67.4} & \underline{63.4}\\
      \rowcolor[RGB]{230,230,230}
      SyLSTM* & \textbf{62.4} & \underline{66.3} & \textbf{64.4}\\
      \hline
    \end{tabular}
    }
    \caption{Offensive Language Target Identification}
    \label{tab:table3}
    \vspace{-0.5cm}
\end{table}

\subsection{Performance on Hate Speech and Offensive Language Dataset}
This section presents the performance comparisons between our model and the baselines for this multi-class classification problem. 
The task presented by the dataset complies with our main objective of integrating syntactic dependencies in a neural network model to differentiate between offensive language and hate speech more efficiently. 
The tweets are classified in one of three categories: hate speech (\textbf{HATE}), offensive language (\textbf{OFF}), and none (\textbf{NONE}). 
The Linear-SVM and the neural network baselines produce very similar results, all of which are significantly better than chance (see Table \ref{tab:table4}). 
The \emph{SyLSTM} (in gray) significantly outperforms all the~baselines. 
\begin{table}[hbt!]
    \centering
    \resizebox{0.9\columnwidth}{!}{
    \begin{tabular}[width=\textwidth]{l l l l}
    \hline
     \textbf{System} & \textbf{Precision} & \textbf{Recall} & \textbf{F1-score} \\
     \hline
      All HATE & 0.2 & 6.1 & 0.4 \\
      All OFF & 3.1 & 16.9 & 5.3 \\
      All NONE & 58.8 & 77.2 & 66.7 \\
    \hline
      SVM & 84.9 & 90.1 & 88.2\\
      BiLSTM & 90.3 & 90.2 & 90.3\\
      BERT & \underline{91.2} & 90.4 & 91.0\\
    \hline
      \rowcolor[RGB]{230,230,230}
      SyLSTM & 90.5 & \underline{91.4} & \underline{91.4}\\
      \rowcolor[RGB]{230,230,230}
      SyLSTM* & \textbf{92.3} & \textbf{92.8} & \textbf{92.7}\\
      \hline
    \end{tabular}
    }
    \caption{Hate Speech and Offensive Language Dataset}
    \label{tab:table4}
    \vspace{-0.5cm}
\end{table}

\subsection{Discussion}
The two-channel BiLSTM and the BERT model discussed in this paper act as strong syntax-agnostic baselines for this study. 
The aforementioned results indicate the superiority of the \emph{SyLSTM} over such approaches. 
The inability of existing dependency parsers to generate highly accurate dependency trees for a tweet may seem like a severe problem. 
However, 
since the dependency tree has been transformed to accommodate inverse dependency edges, we find that the resulting undirected graph acts as a single-relational graph where each edge represents a ``dependency". 
The nature of the dependency is addressed by graph convolutions operating over the dynamic LSTM features. 
Hence, the parser only needs to generate congruent copies of the actual dependency tree of the tweet.

\vspace{0.1cm}We tested the utility of enriching the features generated by a BERT encoder using a GCN. 
Existing literature in this field integrates word embeddings learned using a GCN with the BERT model~\cite{lu2020vgcn}. 
In contrast, our experiments dealt with a GCN mounted over a BERT encoder. 
We note that this combination leads to over-parametrization and severe sparsity issues. 
Since BERT models have been shown to learn fairly accurate dependency structures~\cite{clark:2019}, additional importance to dependency grammar over the same encoder network may be~unnecessary.

\section{Conclusion}
\label{conclusion}
In this paper, we present a novel approach called the \emph{SyLSTM} which demonstrates how GCNs can incorporate syntactic information in the deep feature space, leading to state-of-the-art results for fine-grained offensive language detection on Twitter Data.
Our analysis uncovers the Semantic and Syntactic Encoders' complementarity while revealing that the system's performance is largely unaffected for mislabeled dependencies over congruent dependency trees. 
Leveraging the dependency grammar of a tweet provides a practical approach to simulating how humans read such texts. 
Furthermore, the performance results of the \emph{SyLSTM} indicate the robustness of the architecture in generalizing over small datasets. 
The added simplicity of the overall architecture promotes applicability over other NLP tasks. 
The \emph{SyLSTM} can be used as an efficient and scalable solution towards accommodating graph-structured linguistic features into a neural network~model.

\vspace{0.2cm}\noindent {\bf Replication Package.} The replication package for this study is available at \url{https://github.com/dv-fenix/SyLSTM}.

\bibliography{references}
\bibliographystyle{acl_natbib}

\end{document}